\documentclass[10pt,twocolumn,letterpaper]{article}

\usepackage{cvpr}
\usepackage{times}
\usepackage{epsfig}
\usepackage{graphicx}
\usepackage{amsmath}
\usepackage{amssymb}

\usepackage[breaklinks=true,bookmarks=false]{hyperref}

\cvprfinalcopy 

\def\cvprPaperID{****} 

\begin{document}

\title{Learning Furniture Compatibility with Graph Neural Networks}

\author{\parbox{16cm}{\centering
    {\large Luisa F. Polan\'{i}a, Mauricio Flores, Matthew Nokleby, and Yiran Li}\\
    {\normalsize
    Target Corporation, Sunnyvale, California, USA\\
    Email: \{Luisa.PolaniaCabrera, Mauricio.FloresRios, Matthew.Nokleby, Yiran.Li\}@target.com\\}}
}

\maketitle


\begin{abstract}
We propose a graph neural network (GNN) approach to the problem of predicting the stylistic compatibility of a set of furniture items from images. While most existing results are based on siamese networks which evaluate pairwise compatibility between items, the proposed GNN architecture exploits relational information among groups of items. We present two GNN models, both of which comprise a deep CNN that extracts a feature representation for each image, a gated recurrent unit (GRU) network that models interactions between the furniture items in a set, and an aggregation function that calculates the compatibility score. In the first model, a generalized contrastive loss function that promotes the generation of clustered embeddings for items belonging to the same furniture set is introduced. Also, in the first model, the edge function between nodes in the GRU and the aggregation function are fixed in order to limit model complexity and allow training on smaller datasets; in the second model, the edge function and aggregation function are learned directly from the data. We demonstrate state-of-the art accuracy for compatibility prediction and ``fill in the blank" tasks on the Bonn and Singapore furniture datasets. We further introduce a new dataset, called the Target Furniture Collections dataset, which contains over 6000 furniture items that have been hand-curated by stylists to make up 1632 compatible sets. We also demonstrate superior prediction accuracy on this dataset.


\end{abstract}

\section{Introduction}
The increasing number of online retailers, with ever-expanding catalogs, has led to increased demand for recommender systems that help users find products that suit their interests. While standard recommender based on collaborative filtering rely on user-item interactions, there has been recent interest in recommender systems that leverage computer vision techniques to estimate {\em visual compatibility} from item images \cite{aggarwal2018learning, polania2019learning}. Even though most of the existing work in this field relates to fashion compatibility \cite{dong2017multi,han2017learning, polania2019learning,rubio2017multi,saha2018learning}, the topic of furniture compatibility has recently gained interest in the vision community \cite{aggarwal2018learning, hu2017visual, pan2017deep}.


Although many features of interest for fashion compatibility, such as as color information, also play an important role in furniture compatibility, there are many features that are unique to the problem of furniture compatibility. One of those features is the scale of objects. For example, dainty objects, such as a coffee table and a settee, tend to look good next to weightier, heavier ones, like a pedestal side table or a sofa. Therefore, the problem of furniture compatibility deserves its own study and that is the motivation of this paper. More precisely, given a collection of images of furniture items, we seek a model that predicts the stylistic compatibility of the items in the collection. 

Most approaches to the furniture compatibility problem borrow ideas from particular object retrieval and face verification, training a siamese convolutional neural network (CNN) with triplet or contrastive loss \cite{aggarwal2018learning, hu2017visual, pan2017deep} in order to push compatible items close together in feature space and incompatible items far apart. However, these approaches explicitly capture pairwise relationships between items, whereas set compatibility may depend on relational information among multiple items in the set.

 


In order to capture more complex relational information, we propose a graph-based model for compatibility. We represent each furniture set as a fully-connected graph, in which each node represents a furniture item and each edge represents the interaction between two items. We define a graph neural network (GNN) model that aggregates features across this graph to generate item representations that consider jointly the entire set. Instead of considering pairwise comparisons in isolation, our model therefore can capture interactions among multiple items.


Following the existing literature on GNNs \cite{hamilton2017inductive,kipf2016semi}, our model works by first extracting features from each image in furniture set via a CNN, evolving these features iteratively by passing neighbors' features through a gated recurrence unit (GRU) at each node, and finally aggregating these features into a joint compatibility score of the set. All weights in the model are shared across nodes, which gives the model the ability to handle furniture sets of different size without modification to the architecture.


We present two GNN models. In the first, we extend the siamese model to multiple inputs and branches via the GNN architecture. In particular, we generalize the contrastive loss to an arbitrary number of items. Instead of minimizing the distance between positive pairs and maximizing the difference between negative pairs, we minimize the distance between each item and the centroid of its furniture set for the case of compatible sets and maximize the distance to the centroid for the case of incompatible sets. We train the end-to-end network, including the CNN and GRU parameters via this generalized loss. This encourages a compatible set to have features that are close together, and different sets, which are presumable not compatible, to have features far apart. The compatibility score of a proposed furniture set is calculated by taking the average distance of item embeddings to their centroid.

In the second model, we draw inspiration from graph attention networks \cite{Veli2017}, which learn scalar weights between connected nodes in order to capture richer notions of graph structure. We allow the edge function, which dictates how nodes aggregate neighboring features before putting them into the GRU, to be learned directly from the data instead of fixed in advance. Further, instead of using the distance to the centroid to measure compatibility, we learn a function---in the form of a fully-connected network---that maps GNN features to a compatibility score. This choice is motivated by deep metric learning approaches \cite{Hu2014, Yi2014}, which compared to linear transformations, such as the Euclidean or Mahalanobis distance, are better at capturing the nonlinear manifold where images usually lie on. Because this second model has more parameters to train from data, it requires larger datasets to train, whereas the first model is suitable for smaller datasets.

We demonstrate the utility of our approach by establishing new state-of-the-art performance for furniture compatibility. We perform experiments on the Bonn \cite{aggarwal2018learning} and Singapore datasets \cite{Hu2014} and show superior performance on the tasks of predicting compatibility of furniture sets and "filling in the blank" of partial furniture sets. One challenge with these datasets is that they encode style information only via coarse style categories---every item in the style category is considered to be compatible, which may not correspond to real-world notions of compatibility.

To address this issue, we also introduce a new dataset, which we term the {\em Target Furniture Collections} dataset. This dataset contains over 6000 items that have been organized into 1632 compatible furniture sets. These sets have been chosen by professional furniture stylists, and they encode a richer sense of style compatibility than existing datasets. Instead of supposing that items are compatible because they share the same style attribute, with the Target Furniture Collections dataset we suppose that items are compatible because a stylist has put them in the same set. We also show that our GNN model outperforms existing methods on the Target Furniture Collections dataset.


 The main contributions of this work are summarized as
 follows: (1) We
 propose the first furniture compatibility method that uses
 GNNs, which leverages the relational information between items in a furniture set. (2) Two GNN models are proposed. We propose a generalized contrastive loss function to train the first model, which extends the concept of the siamese network to multiple branches. The second model differs from the first model in that it learns both the edge function and the aggregation function that generates the compatibility score, unlike traditional GNN approaches which use predefined functions. (3) We introduce a new furniture compatibility dataset (available at {https://datahub.io/lfpolani/target-furniture-collections-dataset/v/1}).


\section{Related Work}
The past five years have seen substantial interest in addressing the problems of visual compatibility and style classification for both furniture and fashion. In this section, we provide an overview of prior work, and emphasize the main novelties of our work when compared to prior efforts. 
 
Many authors have framed the problem of visual/style compatibility as a metric learning problem, for instance \cite{ he2016learning, mcauley2015image,polania2019learning} for fashion, as well as \cite{aggarwal2018learning,bell2015learning} for furniture. For example, the model in \cite{polania2019learning} consists of two sub-networks, the first sub-network is a siamese network which extracts feature for the input pair of images, while the second sub-network is a deep metric learning network. The whole model is trained end-to-end in order to derive a notion of compatibility between \emph{a pair} of items. However, such model fails at modeling the complex relations among multiple items. The outfit generation process has been modeled as a sequential process through bidirectional LSTMs in \cite{han2017learning}. However, the assumption of a sequence fails at properly modeling either an outfit or a furniture set, where the concept of a fixed order of items does not exist.

Foundational work in GNNs, such as \cite{gori2005new, scarselli2008graph}, paved the way for a number of applications and breakthroughs in recent years, such as the development of graph convolutional networks or graph attention networks in \cite{velivckovic2017graph}, to name a few. Two GNN approaches have been proposed for fashion, \cite{cui2019dressing} and \cite{cucurull2019context}. The first one creates a fashion graph with nodes corresponding to broad categories (e.g. pants), while compatible outfits are directed subgraphs, while the second developed a context-aware model, to account for personal preference and current trends. Since this paper is focused on furniture compatibility, comparisons are performed only with previously proposed visual compatibility algorithms for furniture. 

 
Until recently, the problem of furniture compatibility has received less attention, and to the best of our knowledge, GNN models have not been developed. The authors of \cite{aggarwal2018learning} addressed the task of style compatibility and style classification among $17$ categories using a Siamese network, and also proposed a joint image-text embedding method. Meanwhile, the authors of \cite{hu2017visual} address the style classification problem by creating handcrafted features using standard feature extractors, such as SIFT and HOG, and then applying an SVM classifier. On a related but somewhat different line of work, \cite{ liu2015style,lun2015elements, pan2017deep} have devised models to predict the style compatibility of computer-generated $3$D furniture models.
 
Prior work using Siamese network approaches suffers from an inability to measure similarity in collections that consist of more than two items. While one could assemble collections of multiple items by aggregating pairwise compatibility scores, such approach disregards the complex stylistic interactions that different items may have. Meanwhile, sequential generation artificially introduces a notion of \emph{time dependence} into the model. 
 



\section{Proposed Method}

Given a furniture set, we aim at predicting the compatibility score of the set. To achieve this goal, we propose to represent a furniture set as a graph, where each node represents an item and each edge represents
interaction between two items.

\begin{figure*}[t]
    \centering
    \includegraphics[width=0.98\textwidth]{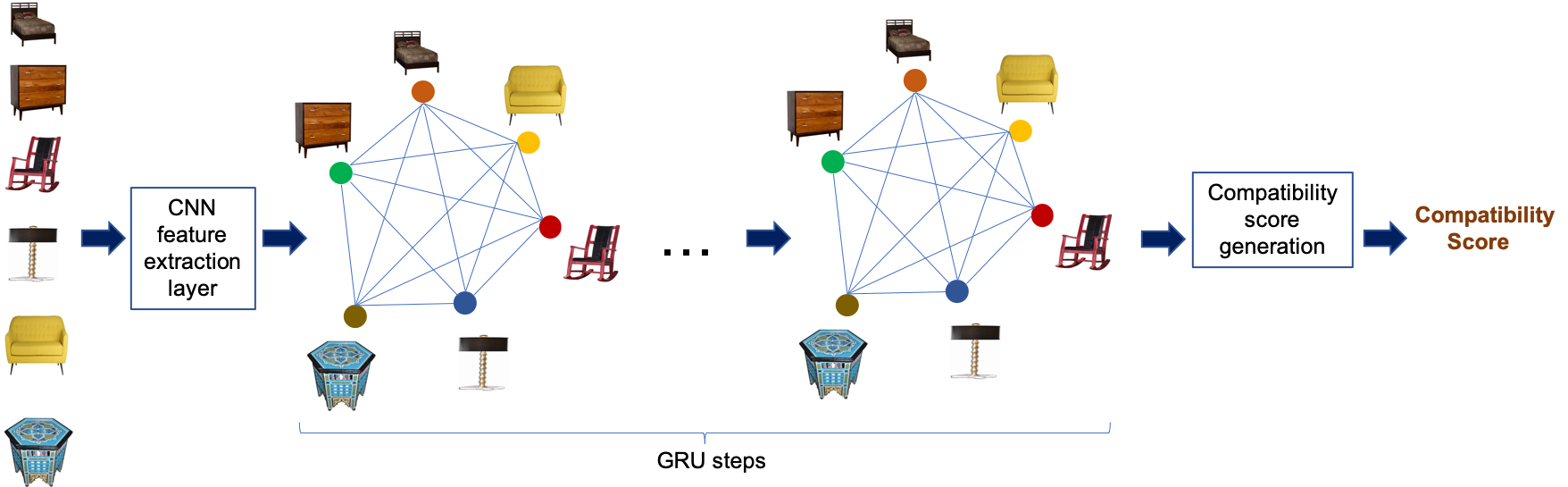}
    \caption{Schematic of the GNN model. CNNs are first used to get the initial  feature  representation  of  the  nodes  and  GRUs  are used to update the state of the nodes during $K$ iterations. The hidden states are then aggregated to generate the compatibility score.}
    \label{fig:schematic}
\end{figure*}

The GNN model can be decomposed into three major steps. The first step learns the initial node representation. The second step models node interactions and updates the hidden state of the nodes by propagating information from the neighbors. The third step calculates the compatibility score. A schematic overview of the GNN model is shown in Figure \ref{fig:schematic}.

In this paper, we consider two variants of the GNN model, referred to as Model I and Model II, which differ in how the compatibility score is calculated and in the definition of the edge function. In Model I, the compatibility score is determined by the average distance between the node states and their centroid and the edge function is predefined. Contrarily, in Model II, the node states are aggregated and further processed to generate the compatibility score and the edge function is learned during training. Those modifications lead to model capacity gains at the  expense  of increasing  the  memory  and  computational  cost.

\subsection{Model I}
Model I can be thought of as a generalization of the siamese model to multiple inputs, with the additional advantage of allowing exchange of information between inputs. Such generalization requires the definition of a new loss function, which extends the idea of mapping item pairs close to each other in the feature space to the entire set.  A detailed description of Model I is provided in this section.

\subsubsection{Network Architecture}
Let $S = \{I_0, I_1, \ldots, I_{N-1}\}$ be the images of the items belonging to an arbitrary furniture set. These images are first mapped to a latent semantic space with a convolutional neural network. This operation is represented as
\begin{equation} \label{eq2}
h_{i}^{0} = \psi(I_i), \hspace{.2cm} i=0, \ldots, N-1,
\end{equation}
where $h_{i}^{0} \in \mathbb{R}^{L}$ and $\psi$ denote the ${L}$-dimensional initial hidden state of node $i$ and the feature extractor operator, \textit{e.g.} AlexNet, respectively. Note that the number of items $N$ may vary across different furniture sets, and  therefore,  each furniture set  has  its  own graph morphology.  

Since the goal is to learn compatibility, the feature representation of an item should also contain information about the items it is compatible with. This is accomplished by iteratively updating the node hidden states with information from the neighbor nodes using a gated recurrent unit (GRU). That is, at every time step $k$, a GRU takes the previous hidden state of the node $h_{i}^{k-1}$ and a message $m_{i}^{k}$ as input and outputs a new hidden state $h_{i}^{k}$.


The message $m_{i}^k \in \mathbb{R}^{M}$ is the result of aggregating the messages from the  node neighbors, and is defined by the aggregation function $\phi(\cdot)$ as

\begin{eqnarray}\label{eq4}
m_{i}^{k}&=&\phi(\{h_{q}^{k}\mid \forall q, q\neq i)\})\\
&=&\frac{1}{N-1}\sum_{q, 
                  q\neq i}
        \text{ReLU}(W_m(h_{q}^{k}  \mathbin\Vert e_{qi}^{k})+b_m),
\end{eqnarray}
where $W_m \in \mathbb{R}^{M\times (L+J)}$ and $b_m \in \mathbb{R}^M$ are trainable parameters of the model and $e_{qi}^k\in \mathbb{R}^J$ is the edge feature between nodes $q$ and $i$, which is calculated with the edge function  $\upsilon(\cdot)$ as  $e_{qi}^k = \upsilon(h_{q}^{k}, h_{i}^{k})$. Note that the neighbors are all the other nodes since the graph is complete.

Model I uses the absolute difference function as the edge function, which is defined as 
$\upsilon(h_{q}^{k}, h_{i}^{k})=|h_{q}^k-h_{i}^k|$, where $|\cdot|$ denotes element-wise absolute value. Note that the absolute difference function is symmetric, \textit{i.e.} $\upsilon(h_{q}^{k}, h_{i}^{k})=\upsilon(h_{i}^{k}, h_{q}^{k})$. The motivation for the choice of the absolute difference function is that it provides information about the distance between two connecting nodes in the feature representation space.


After $K$ GRU steps, the compatibility score generation layer, takes the hidden states $h_{0}^{K}, \ldots, h_{N-1}^{K}$ as input, applies batch normalization, averages their distance to the centroid $c$, and passes the average through a sigmoid function, which maps onto the interval $[0, 1]$ to generate the compatibility score. More formally, the compatibility score $s$ is calculated as

\begin{equation} \label{score_model1}
s = \sigma\left(\frac{1}{N}\sum_{i=1}^N\|h_i^K-c\|_2^2\right),
\end{equation}
where $\sigma(\cdot)$ denotes the sigmoid function.

\subsubsection{Loss Function}

The contrastive loss has been extensively used in the context of siamese networks for learning an embedded feature space where similar pairs are closer
to each other and dissimilar pairs are distant from each other. However, one limitation of the contrastive loss is that it is based only on feature pairs. We propose a generalized version of the contrastive loss that promotes the item embeddings of a compatible furniture set to cluster tightly together while the item embeddings of an incompatible furniture set are pushed away from their corresponding centroid. Let $c$ denote the centroid of the hidden states of the nodes at step $K$, then the generalized contrastive loss for a training instance takes the form

\begin{eqnarray}\label{eq6.18}
L&=&\frac{1}{N}\sum_{i=1}^N\left(yd_{i}^2+(1-y)\text{max}(0, m^2-d_{i}^2)\right)  \nonumber\\
d_{i}&=&\|h_i^K-c\|_2,
\end{eqnarray}
where $y \in \{0,1\}$  is the label with 1 and 0 denoting a compatible and an incompatible furniture set, respectively. The first term of the loss function penalizes compatible furniture sets whose node representations are not tightly clustered around their centroid while the second term penalizes incompatible furniture sets whose node representations are closer than a margin $m$ from their centroid.



\subsection{Model II}
In Model II, the edge function and the aggregation function that generates the compatibility score are learned instead of predefined. Details of Model II are described in this section. 

\subsubsection{Network Architecture}
All the stages before the compatibility score generation are the same as in Model I. That is, CNNs are used to get the initial feature representation of the nodes, and GRUs are used to update the state of the nodes during $K$ iterations, which results in $h_{0}^{K}, \ldots, h_{N-1}^{K}$ hidden states.


Additional steps are added to the network in order to generate the compatibility score.  First, the hidden states are averaged, $1/N\sum_{i=1}^N h_{i}^{K+1}$, and then passed through a a multilayer perceptron (MLP) with ReLU activation and with parameters $\Theta^{(0)}, \ldots, \Theta^{(Q-1)}$, where $\Theta^{(i)}$ are the parameters of the $i$th layer and $Q$ is the number of layers. The last layer outputs the compatibility score $s$, which is normalized in the range $[0, 1]$ through a sigmoid function.


In Model II, the parameters that define the edge function are learned. That is, the edge function connecting node $h_i$ with node $h_j$ is defined as  
\begin{equation} \label{eq2}
\upsilon(h_{i}^{k}, h_{j}^{k})=\text{ReLU}(W_e(h_{i}^{k}\mathbin\Vert h_{j}^{k})+b_e),
\end{equation}
where $W_e \in \mathbb{R}^{J\times 2L}$ and $b_e\in \mathbb{R}^J$ are parameters of the model. Note that unlike traditional graph attention networks \cite{Veli2017}, which only learn scalar weights between connected nodes, the proposed approach learns the edge function.

The binary cross-entropy loss is used as the loss function for training Model II.

\section{Experiments}
Here we detail the datasets used for training and testing our model, describe the experimental settings, and compare the performance of our proposed method to the state of the art.



\begin{figure*}[t]
    \centering
    \includegraphics[width=0.7\textwidth]{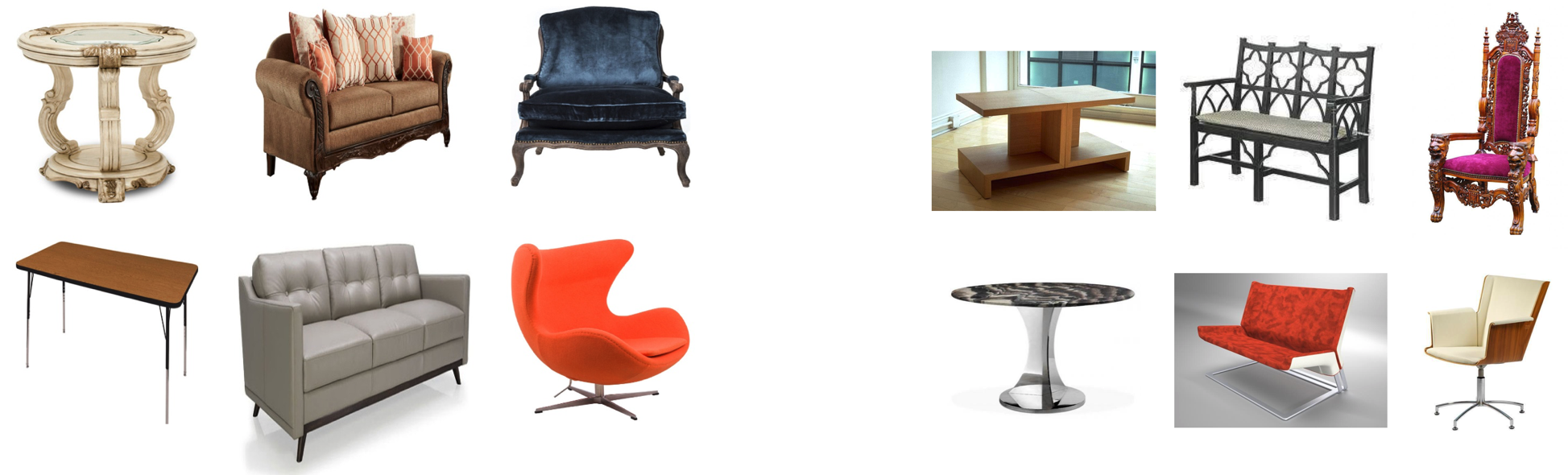}
    \caption{Side by side comparison between the Bonn Furniture dataset (left figures), and the Singapore Furniture dataset (right figures). (a) For Bonn, we see a Victorian style (top) vs. a Modern style (bottom). (b) For the Singapore set, we see a Gothic style (top) vs. a Modernist style (bottom). The Singapore dataset is generally harder to learn from, as images have a mixture between solid-white, realistic and non-realistic backgrounds.}
    \label{fig:bonn}
\end{figure*}

\begin{figure}[t]
\begin{center}
   \includegraphics[scale=0.19]{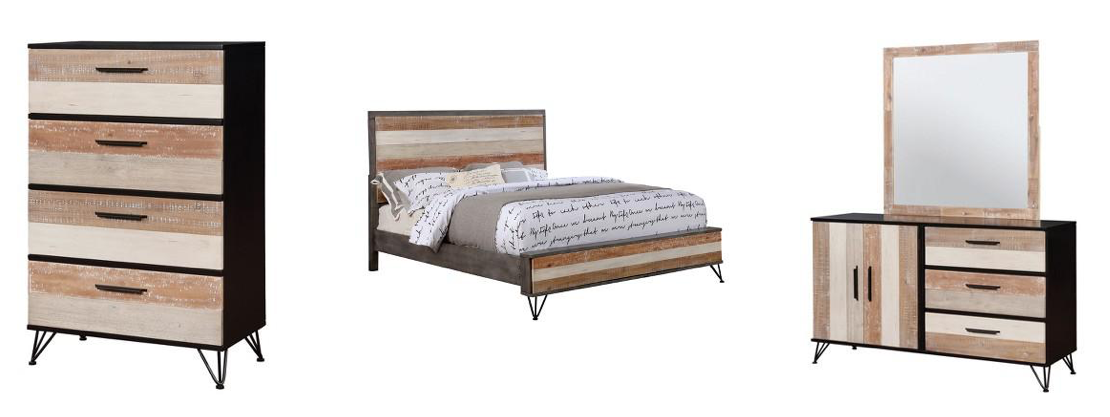}
\end{center}
   \caption{Sample collection from the Target Furniture Collections dataset. The furniture pieces match not only in style, but also color, material and overall appearance.}
\label{fig:}
\end{figure}

\subsection{Furniture Datasets}
Our experiments make use of three datasets, the Bonn Furniture Styles Dataset \cite{aggarwal2018learning}, the Singapore dataset \cite{hu2017visual}, as well as the Target Furniture Collections dataset obtained from the \textit{Target} product catalog. 
\subsection{Singapore Furniture Dataset}
The Singapore Furniture Dataset is the first furniture dataset specifically for furniture style analysis \cite{hu2017visual}. It contains approximately $3000$ images, which are collected from online search tools (Google), social media (Flickr) and the ImageNet dataset. Images are divided into 6 categories: bed($263$), cabinet ($569$), chair ($529$), couch ($391$), table ($429$), and others (774). We adopt the first five categories for consistency within each class. Each image belongs to one of the 16 classes of furniture styles such as American style, Baroque style, etc, and each style contains at least 130 images. Note that some images in this dataset have non-white background and are gathered under realistic scenes. This introduces noises and brings some difficulty in accurately learning and predicting furniture styles.
\subsubsection{Bonn Furniture Styles Dataset}

The Bonn Furniture Styles Dataset consists of approximately $90,000$ furniture images, obtained from the website Houzz.com, which specializes on furniture and interior design. The images span the six most common furniture categories  in  the website,  namely lamps  ($32403$), chairs ($22247$), dressers ($16885$), tables ($8183$), beds ($6594$) and sofas ($4080$). Each image presents the item in a white background and is labelled with one of 17 styles, such as modern, mid-century or Victorian. See Figure \ref{fig:bonn} for a side-by-side comparison of styles. The authors of \cite{aggarwal2018learning} made the dataset available for research and commercial purposes. 

\subsubsection{Target Furniture Collections Dataset}

The Target Furniture Collections dataset contains approximately $6550$ furniture images. The  images  span a wide variety of categories, including $1607$ tables,  $702$ chairs, $406$ dressers, $410$ beds, $350$ sofas, $233$ nightstands, $220$ stools, $154$ benches and over $10$ other categories (such as desks, headboards, drawers, cabinets and mirrors) with a smaller number of items. These items have  been arranged into $1632$ compatible collections by home d\'ecor specialists. These collections vary in size, from $2$ up to $20$ items (though $97$\% of collections contain $8$ items or less). While most collections are complementary in nature, we allow our definition of collection to include any number of compatible items even if a category appears more than once. For example, an office-style collection may include $20$ slightly different office chairs. The dataset is released with a default resolution of $400 \times 400$ pixels.


While this dataset is smaller in size compared to the Bonn dataset, we contend that this dataset provides a richer notion of overall compatibility among furniture sets. While the Bonn dataset classifies furniture pieces across multiple styles, the fact that two items have the same style does not necessarily imply compatibility. The Target Furniture Collections dataset assembles furniture pieces into sets that are compatible not only in style, but also in color, and oftentimes material and texture as well.

\subsection{Compatibility Prediction Task}
For this task, we use our GNN model to predict whether a set of furniture items are compatible. For the Bonn Furniture dataset, we suppose a set of items are compatible if and only if they have the same style attribute, e.g. all ``baroque'' items are compatible, and all ``baroque'' items are incompatible with all ``modernist'' items.  For the Target Furniture Collections dataset, we suppose that two items are compatible if and only if they belong to the same set. We acknowledge that these assumptions artificially limit the definition of compatible---furniture items across style types or sets may well go well together. However, these assumptions provide unambiguous definitions of compatibility, and we maintain that success according to these definitions indicates the extent to which a recommendation model has learned meaningful style attributes.

For each test set, we compute the compatibility score $s$ for Model I and II. We report the area under the ROC curve (AUC).


\subsection{Fill-in-the-blank Task}
The fill-in-the-blank (FITB) task consists of choosing, among a set of possible choices, the item that best completes a furniture set. This is a common task in real life, \textit{e.g.}, a user wants to choose a console table that matches the rest of his living room furniture.

\begin{figure}[t]
\begin{center}
   \includegraphics[scale=0.155]{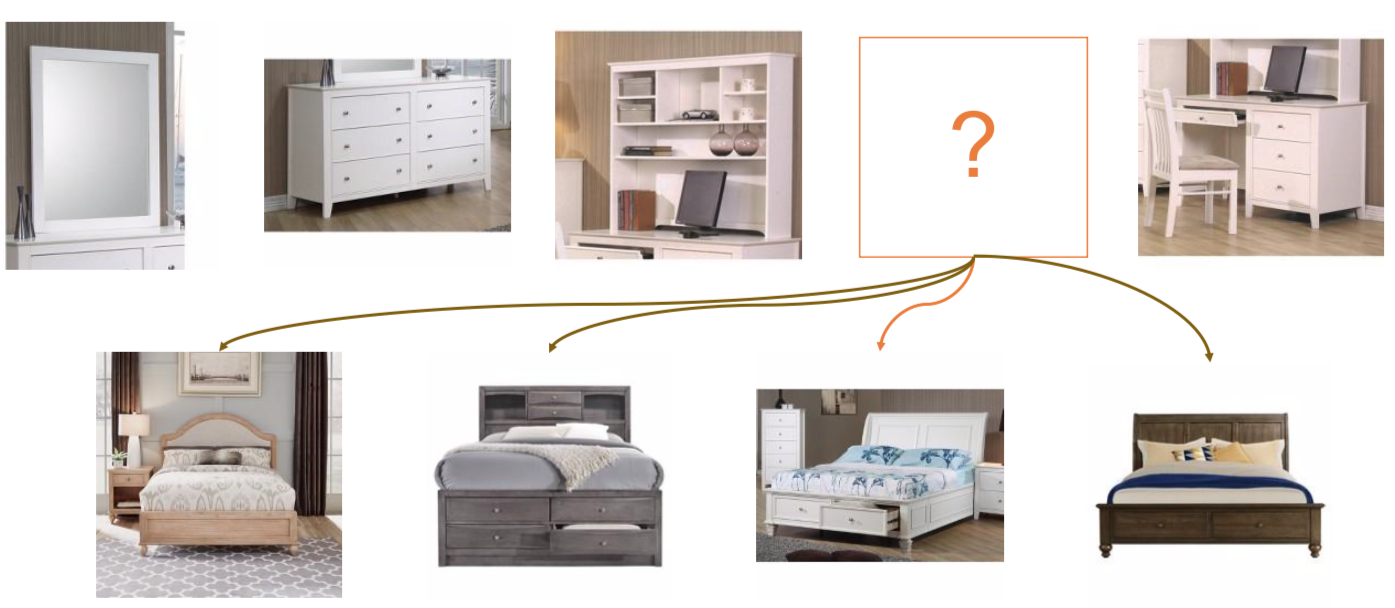}
\end{center}
   \caption{Illustration of the fill-in-the-blank task for furniture compatibility}
\label{fig:FITB}
\end{figure}

The FITB question consists of a set of items that partially form a furniture set and a set of possible choices that includes the correct answer. The number of choices is set to 4. For each dataset, the correct sets correspond to the testing sets. An item is randomly selected from each testing set and replaced with a blank. Fig. \ref{fig:FITB} illustrates an example of a FITB question, where the top row refers to the partial furniture set and the bottom row are the choices for completing the set. 

For the Target Furniture Collections dataset, the incorrect answers are randomly selected items from the same category as the correct answer. For the Bonn Furniture dataset and the Singapore dataset, the incorrect answers are randomly selected items from the same category, but different style, as the correct answer. For example, if the correct answer is a midcentury table, then the incorrect answers are randomly chosen from the pool of tables with style different from midcentury.

This task is addressed by forming all the possible sets between the partial set and the item choices, running the sets through the GNN model and selecting the item that produces the set with the highest score. The performance metric used for this task is the accuracy in choosing the
correct item. Given that the number of choices is 4, the accuracy for a random guess is 25\%.

\subsection{Baseline Experiments and Comparative Results}

Our experiments are compared with results by Aggarwal \textit{et. al.} \cite{aggarwal2018learning}. For comparison purposes, the authors in \cite{aggarwal2018learning} provided us with the siamese models they trained on the Bonn Furniture dataset and the Singapore Furniture dataset and with code to replicate training. For the Target Furniture Collections dataset, we trained the siamese network using the code provided by the authors. The siamese network uses two identical pretrained CNN bases from a truncated version of GoogLeNet for two image inputs in parallel, connected with a fully-connected layer on top to minimize distance of item embeddings belonging to the same style and pushing away item embeddings with different style. 

The CNN used in our experiments to attain the initial node representation is AlexNet with the last layer removed. Therefore, the dimensionality of the node features is $L=4096$. The number of GRU steps, $K$, and the number of layers of the MLP $Q$ are set to 3. The dimension of the messages $M$ and the edge vectors $J$ is set to 4096. The CNN is initialized with AlexNet pre-trained on ImageNet and the rest of the GNN weights are initialized with the Xavier method~\cite{Gloro10}. The first 3 convolutional layers of the CNN are kept frozen during training.

\subsubsection{Data Generation}
For the Bonn Furniture dataset, we use the same data partitions as in \cite{aggarwal2018learning}. They split the dataset along individual furniture items according  to  a  68:12:20 ratio for training, validation and testing. To train our GNN model, we arrange the training set into positive ensembles of variable length, by randomly sampling furniture from the same style category, and negative ensembles of variable length by sampling furniture items from different style categories. The length of the ensembles is randomly selected from the interval $[3, 6]$. The resulting number of positive samples for training, validation and testing is 100K, 13K, and 20K, respectively. The samples generated are balanced, therefore, the number of negative samples is the same as the number of positive samples for each partition.

Similarly, the authors in \cite{aggarwal2018learning} split the  Singapore Furniture dataset according to a 75:25 ratio for training and testing. We use their same testing partition and split their training partition, using 90\% for training and 10\% for validation. We follow the same procedure as with the Bonn Furniture dataset to generate balanced positive and negative furniture sets, with the difference that we fix the set length to 5 to evaluate the performance of the proposed models on furniture sets of fixed size.  The resulting number of positive samples for training, validation and testing is 24K, 2K, 4.8K, respectively.

For the Target Furniture Collections dataset, we split along {\em furniture sets} according to a 75:10:15 ratio for training, validation and testing. These sets make up the positive ensembles, and to produce negative ensembles, we sample at random from the individual items across different furniture collections until the number of negative sets is the same as the number of positive sets. Even though this approach does not guarantee that it would lead to true negatives, it is widely used in the literature  with the argument that randomly selecting items should be less compatible than the choice made by experienced stylists \cite{polania2019learning, aggarwal2018learning,bell2015learning}.

For training the siamese network using the Target Furniture Collections dataset, pairs are built by forming all the possible pair combinations between items belonging the same furniture set. Negative pairs are built by randomly sampling items from different furniture sets.


\subsubsection{Training details}
We train the GNN model for 60 epochs using the Adam optimizer with default momentum values $\beta_1=0.9$ and $\beta_2=0.999$. Hyper-parameters are chosen via cross-validation, which results in a base learning rate of $lr_1=4\times 10^{-6}$ and $lr_2=4\times 10^{-5}$ for the CNN and the rest of the GNN, respectively, and margin $m=50$. For each task and dataset, the criteria for choosing the model was to select the model with the highest AUC in the validation set. The batch size is set to 64 and 32 training samples for Model I and II, respectively.

\begin{table}[]
    \centering
    \begin{tabular}{|c|c|c|c|}
         \hline
         &\shortstack{Bonn\\ (AUC)}
& \shortstack{Target Furniture \\Collections \\ (AUC)}
& \shortstack{Singapore \\ (AUC)}
\\
          \hline
         \shortstack{ Siamese} & 0.865 & 0.907 & 0.945  \\
          \hline
         Model I &0.881 & 0.953  &0.982  \\
          \hline
         Model II &0.897 & 0.931 & 0.989\\
          \hline
    \end{tabular}
    \caption{Comparison of the proposed models with the siamese model using the AUC as performance metric for the compatibility prediction task.}
    \label{tab:compatibility.results}
\end{table}

\begin{figure*}[t]
    \centering
    \includegraphics[width=\textwidth]{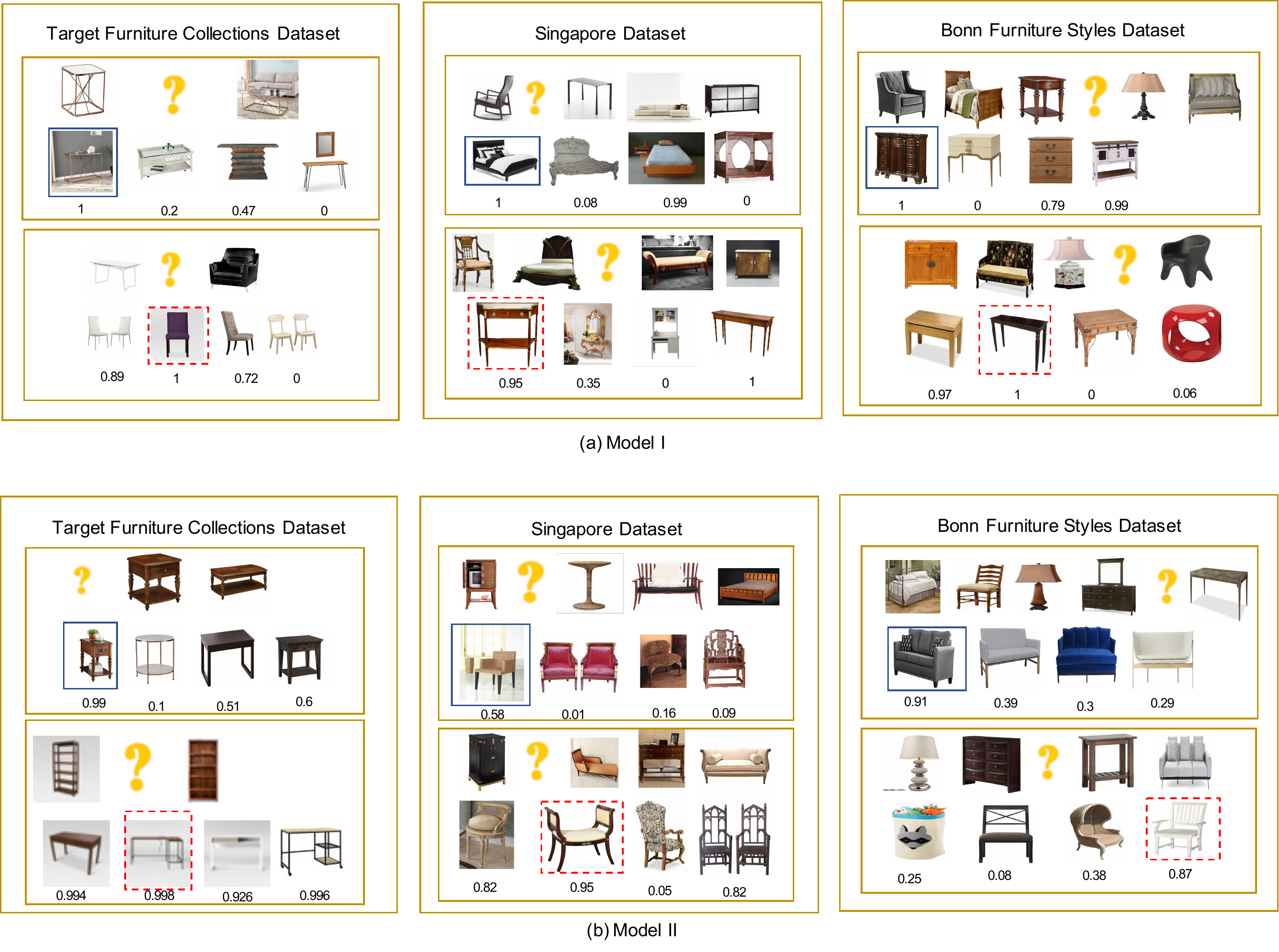}
    
    \caption{Visual results of the FITB tasks for each dataset, using both Model I (a) and Model II (b). For each task, the upper row corresponds to items already in the set, while four options are available in the lower row, alongside with a score for the overall set they form.}
    \label{fig:visual_results}
    \vspace{5pt}
\end{figure*}

\subsubsection{Results for Compatibility Prediction}
Since siamese models predict compatibility based on input pairs but the proposed models predict the compatibility of the entire furniture set, we use the standard approach proposed by previous works on fashion compatibility to compare the two models \cite{ cui2019dressing,han2017learning}. The approach consists of averaging pair-wise compatibility of all pairs in the set.



The comparison between the proposed models and the siamese model for the furniture compatibility prediction task is shown in Table \ref{tab:compatibility.results}. The metrics for both Model I and Model II are higher than for the siamese model, which suggests that learning the relational information between items through a GNN is beneficial. Also, note that if all the possible pairs within the training furniture set were extracted, the result would be 848K pairs, which is much smaller than the 2.2 million training pairs used to train the siamese model in \cite{aggarwal2018learning}. 

Model II outperforms Model I on both the Bonn Furniture dataset and the Singapore dataset, which suggests that learning the functions instead of using predefined functions is beneficial for performance. 


\begin{table}[]
    \centering
    \begin{tabular}{|c|c|c|c|}
         \hline
         &\shortstack{Bonn\\ (ACC)}
& \shortstack{Target Furniture \\Collections \\ (ACC)}
& \shortstack{Singapore \\ (ACC)}
\\
          \hline
         \shortstack{ Siamese} & 0.559  &0.727  & 0.708  \\
          \hline
         Model I & 0.578 & 0.782  & 0.724  \\
          \hline
         \shortstack{Model I\\ (pairwise)} & 0.564 &  0.762 &  0.672 \\ 
          \hline
         Model II & 0.601 & 0.819 & 0.774 \\
          \hline
 \shortstack{Model II\\ (pairwise)} & 0.556 & 0.772  &0.728   \\
          \hline
    \end{tabular}
    \caption{Comparison of the proposed models with the siamese model for the fill-in-the-blank task. Results also include comparisson with the GNN model applied in a pairwise fashion.}
    \label{tab:fitb.results}
\end{table}

\subsubsection{Results for the Fill-in-the-blank task}

The experimental results for the FITB task are shown in Table \ref{tab:fitb.results}. Results of the GNN model applied in a pairwise fashion are also included to further evaluate how the GNN model benefits from exploiting the relational information between items in a set. This means that the input of the GNN model is restricted to a pair of images and results, as in the case of the siamese model, are reported for the averaged pair-wise compatibility of all pairs in the set. The results of the GNN model using both pairs and the furniture set as input outperform the results of the siamese model. It is worth noting that the accuracy of the GNN using the whole set as input is higher than its pairwise counterpart, which validates the hypothesis that exploiting relational information across all items leads to performance improvements.


\subsubsection{Visual Evaluation}   
Figure \ref{fig:visual_results} shows qualitative results using Model I and II, for the fill-in-the-blank task. Each dataset has its own column. For each furniture set, the ground-truth is the left-most item (marked in blue when guessed correctly). When the highest-scoring item does not match the ground truth, the item is marked with a dashed red-box instead. For each model and dataset pair, we selected one right and one wrong prediction for visualization purposes. The compatibility scores of Model I shown in Fig. \ref{fig:visual_results} are normalized between $[0, 1]$ and are inversely proportional to the average distance from the centroid.

Note that the FITB task is particularly challenging because the options are from the same category as the ground truth. Nevertheless, in many cases, the predictions with the highest scores tend to be visually compatible even when the ground truth is not matched.




\section{Conclusions}
In this paper, it was shown that the problem of furniture compatibility can be properly addressed with GNNs in order to leverage the relational information between items belonging to a furniture set. 

Two GNN models were introduced in this paper. The first model was trained with a generalized contrastive loss and the second model, of higher capacity, learns the edge function and the aggregation function that generates the compatibility score.

Experiments on the Bonn Furniture dataset and the Singapore dataset, datasets where compatibility is fully determined by style, and the Target Furniture Collections dataset, a dataset where compatibility is determined by home d\'ecor specialists, show that the proposed models outperform previous methods in the tasks of furniture compatibility prediction and fill-in-the-blank.

{\small
\bibliographystyle{ieee_fullname}
\bibliography{egbib}
}

\end{document}